# Temporal Analysis of World Disaster Risk: A Machine Learning Approach to Cluster Dynamics

Christian Mulomba Mukendi,
Department of Advganced Convergence,
Handong Global University
Pohang, South Korea
mulombachristian@handong.ac.kr

Hyebong Choi,
School of Global Entrepreneurship and
Information Communication Technology,
Handong Global University
Pohang, South Korea
hbchoi@handong.edu

*Abstract*— The evaluation of the impact of actions undertaken is essential in management. This paper assesses the impact of efforts considered to mitigate risk and create safe environments on a global scale. We measure this impact by looking at the probability of improvement over a specific short period of time. Using the World Risk Index, we conduct a temporal analysis of global disaster risk dynamics from 2011 to 2021. This temporal exploration through the lens of the World Risk Index provides insights into the complex dynamics of disaster risk. We found that, despite sustained efforts, the global landscape remains divided into two main clusters: high susceptibility and moderate susceptibility, regardless of geographical location. This clustering was achieved using a semi-supervised approach through the Label Spreading algorithm, with 98% accuracy. We also found that the prediction of clusters achieved through supervised learning on the period considered in this study (one, three, and five years) showed that the Logistic regression (almost 99% at each stage) performed better than other classifiers. This suggests that the current policies and mechanisms are not effective in helping countries move from a hazardous position to a safer one during the period considered. In fact, statistical projections using a scenario analysis indicate that there is only a 1% chance of such a shift occurring within a five-year timeframe. This sobering reality highlights the need for a paradigm shift. Traditional long-term disaster management strategies are not effective for countries that are highly vulnerable. Our findings indicate the need for an innovative approach that is tailored to the specific vulnerabilities of these nations. As the threat of vulnerability persists, our research calls for the development of new strategies that can effectively address the ongoing challenges of disaster risk management.

## I. Introduction

Disaster risk is a complex and dynamic phenomenon caused by the interplay of natural hazards, socio-economic conditions, governance, and conflict. Understanding the multifaceted nature of disaster risk is essential for developing effective strategies to manage it [1]. Extensive research has been conducted on disaster risk, hazards, and their implications resulting in the establishment of several indicators capable to explain and quantify their significance and possible impact. For instance, in [2], an analysis of popular disaster risks index and their usability is discussed. They highlighted the importance to select a suitable index at a country level analysis considering the subcomponents of each which could mislead the real interpretation of risk. The frontier in disaster risk science research is discussed in [3] in the light of progress studies based on the context of China. The research of [4], provides a systematic review of the vulnerability over time and depicts the existing trend in that specific area. The World Risk Index (WRI) was developed by the Alliance Development Work as a new way to assess disaster risk at the country scale since 2011. The concept considered through this index is not focused on the impact of a disaster, such as mortality or economic losses, but rather on measuring the risk of disasters by considering the exposure of people and infrastructure to hazards, their vulnerability to catastrophe as well as their ability to cope with and recover from disasters. It is a more comprehensive way of assessing disaster risk than traditional methods, which focus on the impact of disasters[2], [5]. While research on this topic endeavors have illuminated various facets of disaster risks, contributing to the strides made in this field, the existing frameworks designed to mitigate the impacts of such risks are often conceived for long-term durations[3]. This constitutes a significant constraint, given the capricious nature of these hazards and their escalating repercussions on human lifestyle. Furthermore, despite the implementation of these frameworks, the persistence of peril persists, thereby augmenting the vulnerability of certain nations, particularly those categorized as the least developed[6]. Consequently, avenues for these nations to ameliorate their circumstances remain severely restricted.

This study introduces an innovative approach to understanding the complex dynamics of global disaster risk using the World Risk Index (WRI). This is achieved by adopting a temporal scenario-based framework that is complemented by machine learning methodologies. The primary objective is to assess the short-term prospects of countries in vulnerable states transitioning to more secure positions. The study has the potential to enrich our understanding of disaster risk dynamics at a global scale, providing valuable insights into the use of advanced computational techniques to strengthen a nation's resilience to disasters. Central to our approach is the stratification of countries into discrete clusters, predicated upon their respective disaster risk profiles. Subsequently, we delve into the likelihood of nations either maintaining their cluster categorization or undergoing shifts to disparate clusters as time progresses. The insights derived from this analysis provide essential guidance for comprehensively assessing the actual impact of existing frameworks aimed at mitigating hazards on a global scale. Furthermore, this analysis deepens our understanding of formulating more proactive disaster risk management strategies. At the heart of this investigation lies the World Risk Report, an annual publication that delves into various facets of disaster risk management. The cornerstone of the report is the World Risk Index, a tool designed to illuminate the probability of severe natural events escalating into full-blown disasters across a diverse range of nations[7].

To achieve this, the remainder of this paper is organized as follows: section II is the methodology, followed by section III which deals with the results and discussion. To conclude, the section IV provides an overall conclusion along with limitations and future studies.

## II. METHODOLOGY

### A. Dataset

The dataset utilized spans eleven years and incorporates data from 181 countries. It encompasses variables such as Region, World Risk Index (WRI), Exposure, Vulnerability, Susceptibility, Lack of Coping Capabilities, Lack of Adaptive Capacities, Year, and categorical indicators for Exposure, WRI, Vulnerability, and Susceptibility [1]. The group of Figures 2 depicts the temporal evolution of the numerical variable over time.

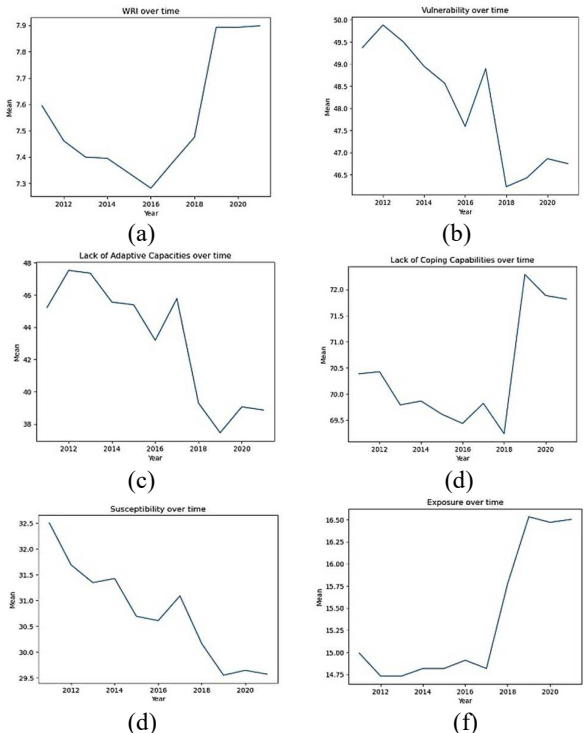

Figures 2: Temporal evolution of WRI and subcomponents

Evidently, there has been a discernible upward trajectory in the Exposure (f) to climate change over the preceding decade, contributing to a notable escalation in the global WRI (a). Likewise, a parallel pattern emerges in the realm of Lack of Coping Capabilities (d), indicating a pressing need for enhanced measures in governance, preparedness, early warning systems, medical infrastructure, as well as social and material security. Despite ongoing efforts in infrastructure development, societal vulnerability to climate change persists. This is evidenced by the existing disparity between vulnerability (b) and susceptibility (d). A similar gap is apparent between the deficiency in coping capabilities (d) and the absence of adaptive capacities (c). This implies that as the magnitude of disaster risks expands, a considerable number of nations are still deficient in the capacity to effectively confront these challenges. Moreover, an even larger cohort struggles to adapt to these events. This intricate scenario could be attributed to multifarious factors, such as economic hardship and other related circumstances.

### B. Outliers detection

The whiskers for WRI and Exposure suggest that there are extreme values which should be considered outliers. These countries are more likely to experience disasters above the normal range, what is not the case for the other variables. By finding the z-score for these two variables with a threshold of 3[8] (*data point that is more than 3 standard deviations away from the mean is considered an outlier*), we can visualize these outliers.

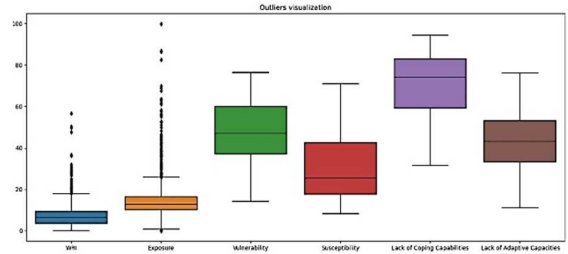

Figure 3: Visualization of outliers in the dataset

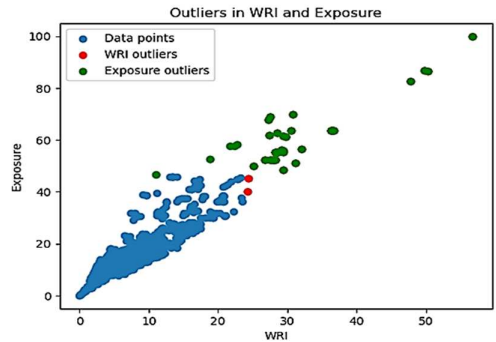

Figure 4: Outliers and inliers in WRI And Exposure

Over time, these countries are: Vanuatu, Tonga, Philippines, Solomon Island, Dominica Republic, Antigua and Barbuda, Brunei and Japan, and there are located in Aisa and Oceania.

### C. Cluster modeling

To unravel the temporal aspects of global disaster risk, our analysis is conducted using a clustering methodology, approach involving the grouping countries into clusters based on their unique disaster risk characteristics as they evolve over time. Thus, the unsupervised modeling was achieved using the KMeans clustering algorithm considering the limited size of the dataset. It is especially valuable when dealing with unlabeled data necessitating the identification of patterns or groupings within the dataset[9] and was employed to cluster countries according to shared characteristics across time predicated on their inherent resemblances. To improve the clustering since they were outliers in the dataset that needed to be considered so that to deepen one's understanding, a semi-supervised learning was achieved using the Label Spreading algorithm[10] which works by first creating a graph of the data, where the nodes of the graph represent the data points and the edges of the graph represent the similarity between the data points and iteratively spreads the labels from the labeled data points to the unlabeled data points, using the edge weights of the graph to determine how likely it is that two data points have the same label. In this research, this algorithm is trained in hiding 50 percent of the labels in the train set, simulating a semi-supervised scenario in which only a small fraction of the data is labeled. The algorithm must learn to classify the unlabeled data based on the labeled data[11].

---

[1]https://www.kaggle.com/datasets/tr1gg3rtrash/globaldisaste-risk-index-time-series-dataset

*D. Supervised learning modeling*

The performance of the Logistic Regression (LR)[12], Decision Tree Classifier (DT)[13], Random Forest Classifier (RF)[14], XGBoost Classifier (XGB)[15] was compared to predict the clusters considering the different period defined. Considering the scenario to analyze, the train and test period was defined as follows:

- For 1 year prediction:
  train = year < 2021 / test = year == 2021
- For 3 years prediction:
  train = year < 2018 / test = year >= 2018
- For 5 years prediction:
  train = year < 2018 / test = year >= 2016

The features considered are: Region, WRI, Exposure, Vulnerability, Susceptibility, Lack of Coping Capabilities, Lack of Adaptive Capacities, Year, Exposure Category, WRI Category, Vulnerability Category, Susceptibility Category and Clusters, the target.

*E. Metrics*

A comprehensive evaluation of the clustering was achieved using the Silhouette Score[16], Calinski-Harabasz Index[17], and Davies-Bouldin Index[18]. These indices offer insights into the quality of clustering outcomes, including cluster separation, compactness, and assignment. For the semi-supervised learning approach and supervised classification task, the AUC[19], confusion matrix[20], accuracy[21].

*F. Scenario analysis*

Scenario analysis stands as a strategic planning and decision-making methodology, involving the thoughtful contemplation of numerous plausible future scenarios. Its purpose lies in assessing potential outcomes, preparing for uncertainties, and making well-informed choices. This technique holds immense value across diverse domains such as business, economics, environmental studies, and risk management. At its core, it revolves around the creation of a collection of distinct scenarios, each portraying a conceivable future condition or a specific set of circumstances. These scenarios typically come to fruition by considering various assumptions pertaining to pivotal variables or factors that hold the power to sway outcomes. By embarking on this exploration of scenarios, decision-makers garner a deeper grasp of the probable risks and opportunities interlinked with the diverse trajectories the future might unfold[22]. Two scenarios were taken into consideration in this research. The first involved evaluating the likelihood of a country remaining within its cluster for the upcoming year, three years, or five years. The second scenario focused on the probability of a country transitioning from its initial cluster to another within the subsequent year, three years, or five years. This approach facilitates the assessment of the efficacy of concerted endeavors aimed at mitigating the adverse repercussions of global hazards within short-term intervals. To operationalize this assessment, a function was developed. This function accepts test data, the current cluster designation, and the predicted cluster designation as inputs. It then filters the data to encompass only countries within the current cluster, calculates the count of countries that have migrated to the predicted cluster, and subsequently computes the probability by dividing the count of migrated countries by the total number of countries in that cluster. The function was instantiated with varying arguments, contingent on the estimation period, to calculate probabilities for diverse scenarios.

*G. Research design*

We followed the process designed in Figure 1 to answer to the scenarios considered in this research.

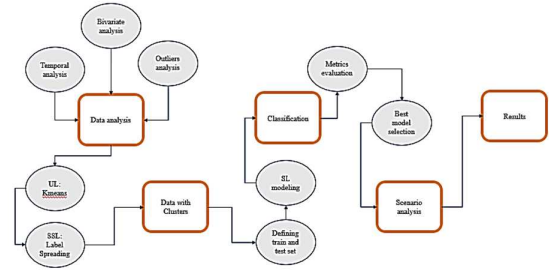

*UL = unsupervised learning, *SSL = semi-supervised learning, *SL = Supervised learning

Figure 1: Diagram of the research design

Upon completing a temporal analysis for each variable to examine their respective trends over time, the dataset undergoes clustering via an unsupervised learning methodology reinforced by a semi-supervised learning modeling. This endeavor results in the partitioning of the data into appropriate formats for classification tasks, where the clusters constitute the target and the variables serve as features. Subsequently, the classifiers aforementioned are used for supervised prediction, and their performance is rigorously assessed to determine the most fitting choice for each distinct scenario. With this optimal algorithm identified, the assessment of probabilities is undertaken, culminating in the articulation of the final outcomes and insights.

III. RESULT AND DISCUSSION

*A. Clustering result*

With a Silhouette Score of 0.46, Calinski-Harabasz Index is 2128.57, and the Davies-Bouldin Index is 0.83, such score

suggests that clustering is good, but not perfect. The data points are not perfectly assigned to their clusters, but they are not poorly assigned either. The clusters are also well-separated, but they are not perfectly compact, this mainly due to the presence of outliers which could not be removed from the dataset since they provide more insight to the overall result. A sample of clusters are provided in the group of Figures 5

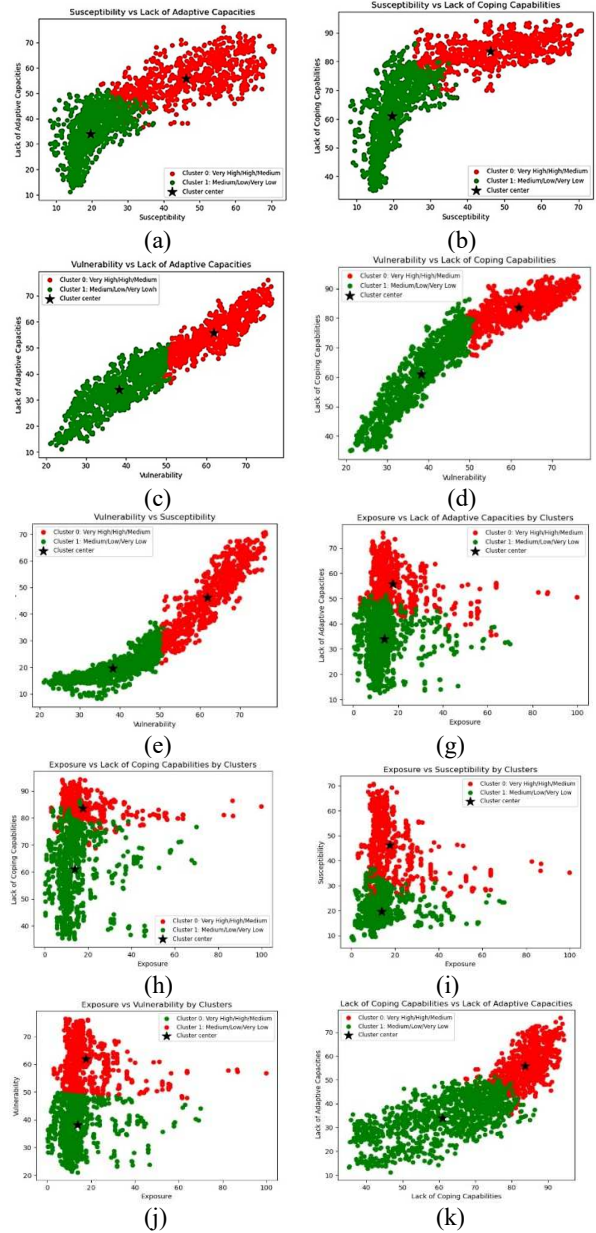

Figure 5: Clusters visualization

It appears that over time, some clusters of countries are well-defined, while others are not. For example, the cluster of countries with low vulnerability and high adaptive capacity (cluster c) is well-defined, meaning that these countries are less likely to be affected by disasters. However, the cluster of countries with high exposure and low coping capacity (cluster h) is not well-defined, meaning that there is a range of countries with different levels of exposure and coping capacity. This suggests that the factors that contribute to disaster risk can change over time. For example, a country that is initially vulnerable to disasters may be able to reduce its vulnerability by investing in adaptive capacity. Conversely, a country that is initially not vulnerable to disasters may become more vulnerable due to factors such as climate change or population growth. Thus, to properly identify the two clusters suggested by the KMeans algorithm, a semi-supervised approach using the Label Spreading algorithm with a K-Nearest Neighbors as kernel was utilized and trained in randomly hiding 50 percents of the label provided by the KMeans. Table I provide the result of the semi-supervised clustering.

TABLE I.    SEMI-SUPERVISED LEARNING PERFORMANCE

| Model | TP | FP | TN | FN | AUC | Accuracy |
|---|---|---|---|---|---|---|
| Label Spreading | 164 | 2 | 2 | 215 | 0.98 | 0.98 |

The 98 percents of accuracy and area under curve alongside with 1% of error in positive probability was achieved and almost 0.9% error on negative probability achieved, suggest that this labeling satisfying compared to the one provided by the unsupervised algoruthm. By reducing the dimensionality of this clustering using the Principal Component Analysis (PCA)[23] with two components, the Figure 6 provide a ploting of the clustering over time with the transduction of the model[24] as option to differenciate the colors of clusters.

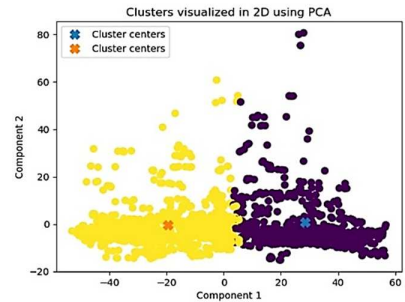

Figure 6: Semi-supervised cluster over time

Figure 6 clearly depicted the separation of clusters over time. The mean coordinate of each cluster was called cluster centers showing how good the clustering was achieved using this approach. The classification task using supervised learning will be achieved on these two classes considering the defined scenario.

*B. Supervised learning classification*

Table II provides a concise summary of the performance attained by each model.

TABLE II.    SUPERVISED LEARNING PERFORMANCE

| Year | Model | TP | FP | TN | FN | AUC | Accuracy |
|---|---|---|---|---|---|---|---|
| One | RF | 67 | 0 | 0 | 114 | 1.00 | 1.00 |
| | DT | 66 | 1 | 0 | 114 | 1.00 | 1.00 |
| | XGB | 67 | 0 | 0 | 114 | 1.00 | 1.00 |
| | LR | 67 | 0 | 0 | 114 | 1.00 | 1.00 |
| Three | RF | 264 | 1 | 2 | 443 | 0.99 | 1.00 |
| | DT | 262 | 3 | 2 | 443 | 0.99 | 0.99 |
| | XGB | 262 | 3 | 2 | 443 | 0.99 | 0.99 |
| | **LR** | **265** | **0** | **1** | **444** | **0.99** | **100** |
| Five | RF | 406 | 6 | 2 | 637 | 0.99 | 0.99 |
| | DT | 409 | 3 | 3 | 636 | 0.99 | 0.99 |
| | XGB | 408 | 4 | 2 | 637 | 0.99 | 0.99 |

| LR | 409 | 3 | 0 | 639 | 0.99 | 1.00 |

In the context of one-year classification, each classifier achieved nearly impeccable predictions, despite the imbalanced classes (67 / 114). The exception was the DT model, which encountered difficulty in accurately classifying False Positives (FP) (1). For three-year classification, the LR exhibited superior performance, boasting reduced error rates for True Negatives (TN) and perfection in False Positives (FP). This positioned it as the optimal choice among the alternatives. Notably, DT and XGB models faced challenges and employed a reduced number of variables compared to their counterparts. Remarkably, the quantity of variables employed by the RF diminished in comparison to its one-year prediction performance. Extending the scope to five-year classification, the LR continued its excellence, with lower error for FP and flawless accuracy for TN. This consistent performance solidified its prominence. Conversely, the RF yielded the least favorable outcome in this scenario. It's worth noting that XGB, while incorporating additional features in its predictions for five years, still fell short of outperforming DT and LR in terms of predictive accuracy. Thus, the prediction achieved by the LR algorithm was considered to evaluate the different probabilities.

*C. Scenario result.*

In accordance with the procedure elucidated in section E of the Methodology, Table III presents the outcomes for each distinct scenario.

TABLE III.  SCENARIO RESULT

| Scenario | Years | | |
|---|---|---|---|
| | *One* | *Three* | *Five* |
| The probability of a country in cluster 0 to remain in this cluster | 1.00 | 1.00 | 0.99 |
| **The probability of a country in cluster 0 shifting to cluster 1** | **0.00** | **0.00** | **0.01** |
| The probability of a country in cluster 1 to remain in this cluster | 1.00 | 1.00 | 1.00 |
| The probability of a country in cluster 1 shifting to cluster 0 | 0.00 | 0.00 | 0.00 |

According to historical data and considering the current global situation with respect to disaster risks evaluate using the WRI and its subcomponents alongside with countries' responses to the hazards, it is highly unlikely (1% of probability) for a country that is highly exposed to disaster risks to shift to a better position in terms of exposure, vulnerability, and susceptibility within the short term (1 to 4 years), regardless of its coping capabilities or adaptive capacities. The impact of improvements in coping capabilities or adaptive capacities may only be observed beyond five years, with a slight possibility of transitioning from a vulnerable position to a better one, although this does not preclude the occurrence of risks. The outcomes of this research contribute significantly to comprehending the extent to which current responses to disaster risks fall short in ensuring global security. Consequently, it becomes imperative to formulate rapid, resilient, and practical solutions that can more effectively counteract these hazards, particularly in ̵nations highly vulnerable and prone to susceptibility.

However, the challenge remains considerable due to the majority of these nations being categorized as least developed countries, implying a lack of resources to address their fundamental needs. Furthermore, the process of establishing robust infrastructures akin to those present in developed countries is a prolonged endeavor, while disaster risks persist as constant and unpredictable threats. The consensus within the scientific community emphasizes the vital necessity of coping capabilities to confront these hazards[7]. This consensus underpins the formulation of various globally recognized roadmaps. Nonetheless, given the nature and magnitude of these hazards, relying solely on long-term preparations proves inadequate for least developed countries. These nations remain exposed, their vulnerability and susceptibility deepening over time. This predicament necessitates a paradigm shift, one that addresses these disasters within these countries through a short-term lens. This approach hinges on addressing specific risks that hold the highest likelihood of occurrence in distinct regions or nations. Consequently, reimagining disaster risk management in alignment with a country's profile and susceptibility emerges as an essential and imperative challenge demanding resolution.

IV. CONCLUSION, LIMITATIONS AND FUTURE STUDIES

Disaster risks are universally critical issues with the potential to cause substantial disruption, transcending geographical boundaries and resisting existing mitigation efforts. In this study, our inquiry delves into the effectiveness of concerted attempts to address this issue over a span of time across the globe. Upon careful analysis of disasters risks using the world risk index and its subcomponents, from the results obtained it was evident that over an 11-year timeframe, the world's composition has bifurcated into two primary clusters, demarcating regions of higher vulnerability from those less susceptible. This trend persists regardless of geographical location and the existing classification categories of the World Risk Index (e.g., medium risk). Additionally, it's apparent that transitioning a country from a precarious position to a safer one within a short-term timeframe is an improbable feat. As such, advocating for a new paradigm of disaster risk management tailored on a short-term based and suitable to susceptible nations becomes a compelling necessity. However, certain limitations underscore this analysis. Even within the delineated clusters, certain countries might not conform to the general cluster characteristics established by algorithms. Deeper exploration at the country level is imperative to comprehend the preparatory measures each nation is undertaking to confront these hazards. This inquiry should also encompass assessing the appropriate budgetary considerations aimed at enhancing short-term coping capabilities. The heterogeneity of hazard exposures across countries necessitates varied levels of effort in addressing these risks. Moreover, the world risk index is an aggregate of 27 publicly available indicators. Knowing that all the countries are not exposed to the same risks, further country based or regional based research need to be carried out to better integrate resource to face their inherent hazards. This brings to light the importance of further research in gauging the extent to which vulnerable countries must invest in bolstering their coping capacities to effectively counter short-term risks.

## V. REFERENCES

[1] S. S. Patel, B. McCaul, G. Cáceres, L. E. R. Peters, R. B. Patel, and A. Clark-Ginsberg, "Delivering the promise of the Sendai Framework for Disaster Risk Reduction in fragile and conflict-affected contexts (FCAC): A case study of the NGO GOAL's response to the Syria conflict," *Prog. Disaster Sci.*, vol. 10, p. 100172, Apr. 2021, doi: 10.1016/j.pdisas.2021.100172.

[2] M. Garschagen, D. Doshi, J. Reith, and M. Hagenlocher, "Global patterns of disaster and climate risk—an analysis of the consistency of leading index-based assessments and their results," *Clim. Change*, vol. 169, no. 1–2, p. 11, Nov. 2021, doi: 10.1007/s10584-021-03209-7.

[3] P. Shi *et al.*, "Disaster Risk Science: A Geographical Perspective and a Research Framework," *Int. J. Disaster Risk Sci.*, vol. 11, no. 4, pp. 426–440, Aug. 2020, doi: 10.1007/s13753-020-00296-5.

[4] B. J. Kim, S. Jeong, and J.-B. Chung, "Research trends in vulnerability studies from 2000 to 2019: Findings from a bibliometric analysis," *Int. J. Disaster Risk Reduct.*, vol. 56, p. 102141, Apr. 2021, doi: 10.1016/j.ijdrr.2021.102141.

[5] M. W. A. Ramli, N. E. Alias, Z. Yusop, and S. M. Taib, "Disaster Risk Index: A Review of Local Scale Concept and Methodologies," *IOP Conf. Ser. Earth Environ. Sci.*, vol. 479, no. 1, p. 012023, Jun. 2020, doi: 10.1088/1755-1315/479/1/012023.

[6] N. Kuruppu and R. Willie, "Barriers to reducing climate enhanced disaster risks in Least Developed Country-Small Islands through anticipatory adaptation," *Weather Clim. Extrem.*, vol. 7, pp. 72–83, Mar. 2015, doi: 10.1016/j.wace.2014.06.001.

[7] Dr. Mariya Aleksandrova *et al.*, "World Risk Report 2021," IFVH, 2021. [Online]. Available: https://weltrisikobericht.de/archiv-e-2021/

[8] R.-M. Penninkangas *et al.*, "Low length-for-age Z-score within 1 month after birth predicts hyperdynamic circulation at the age of 21 years in rural Malawi," *Sci. Rep.*, vol. 13, no. 1, p. 10283, Jun. 2023, doi: 10.1038/s41598-023-37269-9.

[9] M. Ahmed, R. Seraj, and S. M. S. Islam, "The k-means Algorithm: A Comprehensive Survey and Performance Evaluation," *Electronics*, vol. 9, no. 8, p. 1295, Aug. 2020, doi: 10.3390/electronics9081295.

[10] V. Sankar, M. Nirmala Devi., and M. Jayakumar., "Data Augmented Hardware Trojan Detection Using Label Spreading Algorithm Based Transductive Learning for Edge Computing-Assisted IoT Devices," *IEEE Access*, vol. 10, pp. 102789–102803, 2022, doi: 10.1109/ACCESS.2022.3209705.

[11] O. R. A. Almanifi, J. K. Ng, and A. P. P. Abdul Majeed, "The Classification of FTIR Plastic Bag Spectra via Label Spreading and Stacking," *MEKATRONIKA*, vol. 3, no. 2, pp. 70–76, Jul. 2021, doi: 10.15282/mekatronika.v3i2.7390.

[12] X. Zou, Y. Hu, Z. Tian, and K. Shen, "Logistic Regression Model Optimization and Case Analysis," in *2019 IEEE 7th International Conference on Computer Science and Network Technology (ICCSNT)*, Dalian, China: IEEE, Oct. 2019, pp. 135–139. doi: 10.1109/ICCSNT47585.2019.8962457.

[13] V. G. Costa and C. E. Pedreira, "Recent advances in decision trees: an updated survey," *Artif. Intell. Rev.*, vol. 56, no. 5, pp. 4765–4800, May 2023, doi: 10.1007/s10462-022-10275-5.

[14] M. Schonlau and R. Y. Zou, "The random forest algorithm for statistical learning," *Stata J. Promot. Commun. Stat. Stata*, vol. 20, no. 1, pp. 3–29, Mar. 2020, doi: 10.1177/1536867X20909688.

[15] G. Abdurrahman and M. Sintawati, "Implementation of xgboost for classification of parkinson's disease," *J. Phys. Conf. Ser.*, vol. 1538, no. 1, p. 012024, May 2020, doi: 10.1088/1742-6596/1538/1/012024.

[16] K. R. Shahapure and C. Nicholas, "Cluster Quality Analysis Using Silhouette Score," in *2020 IEEE 7th International Conference on Data Science and Advanced Analytics (DSAA)*, sydney, Australia: IEEE, Oct. 2020, pp. 747–748. doi: 10.1109/DSAA49011.2020.00096.

[17] X. Wang and Y. Xu, "An improved index for clustering validation based on Silhouette index and Calinski-Harabasz index," *IOP Conf. Ser. Mater. Sci. Eng.*, vol. 569, no. 5, p. 052024, Jul. 2019, doi: 10.1088/1757-899X/569/5/052024.

[18] J. Xiao, J. Lu, and X. Li, "Davies Bouldin Index based hierarchical initialization K-means," *Intell. Data Anal.*, vol. 21, no. 6, pp. 1327–1338, Nov. 2017, doi: 10.3233/IDA-163129.

[19] Sarang Narkhede, "Understanding AUC - ROC Curve." Towards data science, Oct. 08, 2019. [Online]. Available: https://towardsdatascience.com/understanding-auc-roc-curve-68b2303cc9c5

[20] M. Heydarian, T. E. Doyle, and R. Samavi, "MLCM: Multi-Label Confusion Matrix," *IEEE Access*, vol. 10, pp. 19083–19095, 2022, doi: 10.1109/ACCESS.2022.3151048.

[21] G. Zamariola, P. Maurage, O. Luminet, and O. Corneille, "Interoceptive accuracy scores from the heartbeat counting task are problematic: Evidence from simple bivariate correlations," *Biol. Psychol.*, vol. 137, pp. 12–17, Sep. 2018, doi: 10.1016/j.biopsycho.2018.06.006.

[22] J. Kwesi-Buor, D. A. Menachof, and R. Talas, "Scenario analysis and disaster preparedness for port and maritime logistics risk management," *Accid. Anal. Prev.*, vol. 123, pp. 433–447, Feb. 2019, doi: 10.1016/j.aap.2016.07.013.

[23] A. Ibrahim *et al.*, "Water quality modelling using principal component analysis and artificial neural network," *Mar. Pollut. Bull.*, vol. 187, p. 114493, Feb. 2023, doi: 10.1016/j.marpolbul.2022.114493.

[24] M. Chalvidal, T. Serre, and R. VanRullen, "Learning Functional Transduction," 2023, doi: 10.48550/ARXIV.2302.00328.